\documentclass[letterpaper, 10pt, conference]{ieeeconf}      
\IEEEoverridecommandlockouts                              

\usepackage{graphicx} 
\usepackage{times} 
\usepackage{amsmath} 
\usepackage{bm}
\usepackage{ragged2e}
\usepackage{xfrac}
\usepackage{amssymb} 
\usepackage{algorithm}
\usepackage[noend]{algpseudocode}
\usepackage{multirow,color}
\usepackage{cite}
\usepackage{algpseudocode}
\usepackage{varwidth}
\usepackage{subfig}
\usepackage{booktabs}
\usepackage{breqn}
\usepackage{gensymb}
\usepackage[hyphens]{url}
\usepackage[export]{adjustbox}
\usepackage[font=footnotesize]{caption}
\let\labelindent\relax
\usepackage[inline]{enumitem} 
\usepackage{dblfloatfix}
\usepackage{tabularx}
\usepackage{setspace}
\usepackage{textcomp}
\usepackage{multirow}
\usepackage{comment}
\usepackage{xcolor}
\usepackage{nicematrix}

\usepackage{color}
\usepackage{graphicx,tipa}

\makeatletter
\def\BState{\State\hskip-\ALG@thistlm}
\makeatother

\title{\LARGE \bf An FBG-based Stiffness Estimation Sensor for \textit{In-vivo} Diagnostics}

\author{Behnam Moradkhani, Pejman Kheradmand, Harshith Jella, Kent K. Yamamoto,\\ Alireza Tofangchi, Patrick J. Codd, and Yash Chitalia
\thanks{B. Moradkhani, P. Kheradmand, H. Jella, and Y. Chitalia are with the Healthcare Robotics and Telesurgery (HeaRT) Laboratory, Speed School of Engineering, University of Louisville, Louisville, KY, USA.}
\thanks{A. Tofangchi is with Louisville Automation and Robotic Research Institution (LARRI), University of Louisville, Louisville, KY, USA.}
\thanks{K.K. Yamamoto and P.J. Codd are with the Brain Tool Lab, Department of Mechanical Engineering and Materials Science, Duke University, Durham, NC, USA.}
\thanks{{\textit{Corresponding author: Behnam Moradkhani} (behnam.moradkhani@louisville.edu)}}
}
\begin{document}
\maketitle
\begin{abstract}
\textit{In-vivo} tissue stiffness identification can be useful in pulmonary fibrosis diagnostics and minimally invasive tumor identification, among many other applications. In this work, we propose a palpation-based method for tissue stiffness estimation that uses a sensorized beam buckled onto the surface of a tissue. Fiber Bragg Gratings (FBGs) are used in our sensor as a shape-estimation modality to get real-time beam shape, even while the device is not visually monitored. A mechanical model is developed to predict the behavior of a buckling beam and is validated using finite element analysis and bench-top testing with phantom tissue samples (made of PDMS and PA-Gel). Bench-top estimations were conducted and the results were compared with the actual stiffness values. Mean RMSE and standard deviation (from the actual stiffnesses) values of 413.86 KPa and 313.82 KPa were obtained. Estimations for softer samples were relatively closer to the actual values. Ultimately, we used the stiffness sensor within a mock concentric tube robot as a demonstration of \textit{in-vivo} sensor feasibility. Bench-top trials with and without the robot demonstrate the effectiveness of this unique sensing modality in \textit{in-vivo} applications.
\end{abstract}
\section{Introduction}

Idiopathic pulmonary fibrosis (IPF) is the most common form of the idiopathic interstitial pneumonia, considered lethal due to its unknown causes \cite{king2011idiopathic} with a reported average survival of 4 years after diagnosis \cite{zheng2022mortality}. This disease is characterized by excessive scarring, which leads to increased tissue stiffness, lung dysfunction and ultimately death \cite{upagupta2018matrix}. There is a need for diagnosing IPF in its early stages, for timely interventions. From a broader perspective, tumor identification in tight anatomical spaces has several applications, such as in glioblastomas in the brain \cite{svensson2022decreased} and intraductal papillary mucinous neoplasms (IPMN) in the pancreatic duct \cite{igata2023relationships}. Elastography \cite{chang2011measurement} has emerged as a medical imaging modality for mapping the elastic properties and stiffness of soft tissue. However, this technique has several drawbacks including the need for time-consuming procedures, specialized expertise, and expensive equipment. As a result, dependable alternatives that involve measuring tissue stiffness via local probing and indenting would serve an important purpose in  diagnostics.

Numerous generations of indentation instruments have been developed which generally employ ultrasonic \cite{han2003novel, zheng1996ultrasound, jalkanen2010hand}, electromagnetic \cite{lu2009hand}, or mechanical systems \cite{levental2010simple, zodagehand, li2021portable}, each including a load cell for measuring the applied force and a displacement transducer for recording tissue deformation. However, the mechanical indentation setups of these devices are typically either intricate or large, rendering them unsuitable for \textit{in vivo} testing. As a result, researchers have turned their attention to developing state-of-the-art handheld instruments that permit quick tissue probing and stiffness measurements. Li et al. \cite{li2021portable} introduced a novel and straightforward approach for non-invasive \textit{in vivo} measurement of the elastic modulus of soft tissues based on the buckling phenomenon in a long slender bar and Hertz contact theory. They developed a handheld portable device that utilizes this method to measure tissue stiffness (in the form of the tissue's Young's modulus), over a wide range with a short measurement time. The device's versatility in measuring a broad range of stiffness values distinguishes it from other devices used for this purpose. However, the need for visual inspection to capture the buckling angle at the contact point is a potential limitation of the device, making it challenging to use for \textit{in vivo} surgical operations with visual occlusion, for example, in IPF diagnostics. Additionally, the device's lack of steerability makes it invasive. To overcome this issue, an alternative approach could be to incorporate a sensor within a steerable tube robot to estimate the shape of the buckled beam without relying on a camera.

Although Fiber Bragg gratings (FBGs) have been traditionally employed for measuring temperature and strain in machinery, they have recently been utilized in medical robots for shape and force estimation, due to their small size, biocompatibility, and electromagnetic transparency. 
Using FBGs to estimate a substrate's shape typically involves deviating the neutral axis of the sensing assembly from the center-line of the fiber. In one study \cite{liu2015large}, a triangular cross-section assembly was formed by attaching an FBG fiber to two nickel-titanium (NiTi) wires, which was later used in \cite{sefati2016fbg, sefati2019fbg}. In another investigation \cite{araujo2001temperature}, two D-type optical fibers were attached side-by-side along the plane created by the D-shape cladding, taking advantage of the curvature sensitivity of gratings etched on the fibers. Another method was employed in \cite{roesthuis2013three}, where the authors developed a flexible Nitinol needle with three micro-machined grooves along its length for attaching optical fibers. This design displaced the neutral axis of each fiber, allowing for the measurement of bending strain and estimation of the 3D shape of the robotic needle. Additionally, Xu et al. \cite{xu2016curvature} utilized a helically wrapped FBG sensor design to measure torsion, curvature, and force in a concentric tube robot. In \cite{chitalia2020towards}, a novel sensing assembly was described that uses an FBG fiber attached to a spine, and in \cite{sheng2019large}, the authors developed an FBG bending sensor for SMA bending modules for steerable surgical robots. All these cases demonstrate the repeatability, precision, and capability of FBG-based curvature sensors to measure a wide range of curvature values.

In recent years, numerous endeavors have used FBG sensors to assess tissue stiffness. Li et al. presented a novel approach utilizing a tactile sensor array of five FBG fibers to detect tumors through minimally invasive \textit{in situ} tissue palpation \cite{li2018high}. However, the proposed device in this study is relatively bulky and expensive due to the utilization of multiple fibers. Furthermore, the device is limited to distinguishing between soft and hard surfaces without providing any measurements of specific physical properties. Additionally, noteworthy progress has been achieved in utilizing FBG-based indentation techniques for evaluating cartilage stiffness by the authors in \cite{marchi2017microindentation, marchi2019fiberoptic}. The authors have presented an innovative micro-indentation configuration for determining bovine knee cartilage stiffness parameters. In a subsequent study \cite{kalahrodi20192d}, a similar setup was implemented, but with an additional FBG fiber incorporated to account for temperature changes. However, a significant limitation of their proposed apparatus is its incapability to conduct \textit{in vivo} measurements.

In this work, we propose an entirely novel FBG-based stiffness sensing assembly capable of being advanced through the lumen of a robotic endoscope to lightly indent the surface of tissue \textit{in vivo} and measure its stiffness properties. This approach may be useful in mapping the interior stiffness of peripheral tissue for cancerous malformations and identifying candidate tumor sites in tight anatomical spaces. The proposed sensor acts as a buckling beam while it is being indented into the tissue, and its shape is being estimated using the FBG fiber. The amount of indentation made through the surface of the tissue is then used to distinguish between soft and stiff tissue samples.

The remainder of this paper is organized as follows: We first describe the design of our sensor in Section \ref{sec:Mat_And_Met}. Then, a mechanical model of the sensor assembly is presented in Section \ref{sec:Model}, and is validated in simulation and experimentally. Next, we present bench-top primary estimations, install the sensor within a mock concentric tube robot to conduct indentation trials using the proposed sensor toward \textit{in vivo} applications, and present a failure case analysis in Section \ref{sec:experiments_results}. We conclude this paper by offering a comprehensive summary and final insights in Section \ref{sec:conclusions}. 
\section{Materials and Methods} \label{sec:Mat_And_Met}
\subsection{Design and Fabrication} \label{subsec:Des_And_Fab}
\begin{figure}
\vspace{0.1cm}
\centering\includegraphics[width=\linewidth,keepaspectratio]{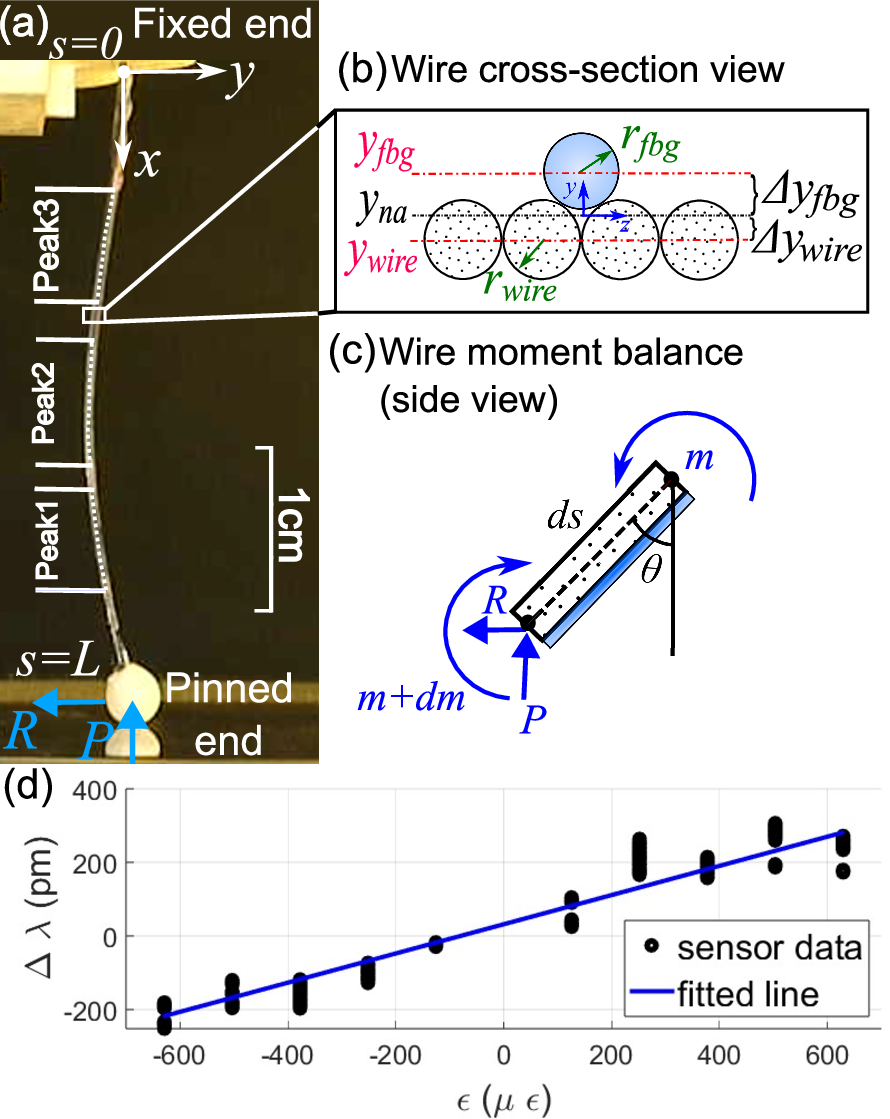}
\caption{(a) Sensor assembly undergoing vertical force application $P$, exhibiting fixed-pinned buckling deformation with key elements, including the reaction force $R$, optical fiber grating regions/peaks, and global coordinates being shown, (b) Cross-section of the sensor assembly used for computing the combined neutral axis, (c) Beam moment balance for a differential section with the length $ds$ of the sensing assembly, (d) validation data for the modified FBG coefficient due to gluing (fitted by a line with the slope of $0.424~\text{pm}/{\mu\epsilon}$).}
\label{fig:modeling_schematic}
\vspace{-0.4cm}
\end{figure}
The sensor assembly (see Fig. \ref{fig:modeling_schematic}(a)) is a composite beam manufactured by attaching an FBG fiber of radius $r_{fbg}=0.115$~mm to a substrate made of four Nickel-Titanium (NiTi) wires of radius $r_{wire}=0.1$~mm attached together in a row (see cross-section view in Fig. \ref{fig:modeling_schematic}(b)). The fiber has five grating elements machined on it at intervals of 8~mm, however, only three are used for the purpose of this paper. A 3D-printed spherical tip (printed with a Projet 2500 Plus MJP, 3D Systems Inc.), is attached to the tip of the composite beam to prevent the beam from perforating the tissue. Furthermore, the spherical contact point can conveniently be modeled as a pin joint such that the beam can be modeled as a fixed-pinned buckling beam. The principal idea behind the sensing mechanism is to advance the beam onto the surface of the tissue until it buckles and use the buckled shape of this composite beam to estimate indentation depth and the exerted force. The mechanical relationship between these parameters was described in \cite{li2021portable}. \textcolor{black}{The post-contact indentation depth of the beam is closely correlated with the elastic modulus of the underlying tissue, which will be made evident in the model proposed in this work.} However, the idea is intuitive: \textit{a beam coming in contact with a hard surface will begin buckling \textcolor{black}{almost} immediately post-contact, whereas a beam coming in contact with softer bodies \textcolor{black}{bends the surface of the tissue inward and} sinks deeper before it begins buckling.} We use this phenomenon for stiffness estimation.

\subsection{Strain-Curvature Relation}
Fiber Bragg gratings are optical fibers with a periodic modulation of the refractive index so that a \textcolor{black}{specific wavelength of light is reflected back from the grated element}. This wavelength is known as the Bragg wavelength of the fiber ($\lambda_b$). Multiple such grating elements can be machined along the length of the fiber, each corresponding to a certain Bragg wavelength. The specific wavelengths reflected back from these gratings are usually visualized as multiple peaks in frequency domain. Therefore, the term ``peak'' is used throughout this paper to refer to these gratings (See Fig. \ref{fig:modeling_schematic}(a) for the locations of the peaks).
The main property of an FBG utilized in continuum robot shape sensing \cite{chitalia2020towards} is as follows: The wavelength depends on the spacing between consecutive gratings in a single element ($\Delta_b$), making it sensitive to axial strain ($\epsilon$) and temperature variations ($\Delta T$). Therefore, the change in Bragg wavelength ($\Delta \lambda_b$) is given as follows 
\begin{equation} \label{eq:Wavelength_Strain_Relation}
    \Delta \lambda_b = k_{\epsilon} \epsilon + S_{T} \Delta T
\end{equation}
The coefficients \textcolor{black}{$k_{\epsilon} = 1.2~\text{pm}/{\mu\epsilon}$} and \textcolor{black}{$S_{T} = 10~\text{pm}/{^{\circ}\text{C}}$} are provided by the manufacturer \textcolor{black}{(Technica Optical Components, Atlanta, United States)}. Here, we assume that temperature stays constant, i.e., $\Delta T = 0$. This may not be the case \textit{in-vivo}, in which case, temperature compensation will be applied in our future work. Additionally, the properties of the fiber seem to have changed due to gluing (using UV glue), and based on the measurements conducted after manufacturing the sensor assembly (showed in Fig. \ref{fig:modeling_schematic}(d)), a new strain coefficient \textcolor{black}{$k'_{\epsilon} = 0.424~\text{pm}/{\mu\epsilon}$} is used instead.

Based on the structure of the sensor assembly and its geometry, the strain at each point along the fiber ($\epsilon_{fbg}$) is related to the curvature along the beam ($\frac{d\theta}{ds}$, which will be elaborated further in Section \ref{sec:Model}) according to the expression:
\begin{equation} \label{eq:Strain_Curvature_Relation}
    \epsilon_{fbg}(s) = \Delta y_{fbg} \frac{d\theta}{ds}
\end{equation}
where the cross-sectional distance between $y_{fbg}$ (FBG fiber's center-line) and $y_{na}$ (neutral axis of the whole assembly) is depicted as $\Delta y_{fbg}$ (See Fig. \ref{fig:modeling_schematic}(b)).
\section{Mechanical Model} \label{sec:Model}
\subsection{Buckling and Post-Buckling Model} \label{subsec:buckling_model}
 For a small arc-length, $ds$, of a beam of total length $L$, the differential moment balance equation is given as follows:
\begin{equation} \label{eq:moment_balance}
    m + dm + Pds\sin{\theta} + Rds\cos{\theta} = m
\end{equation}
Here the moment $m$, forces $P$ and $R$ are indicated in Fig. \ref{fig:modeling_schematic}(c). From the moment curvature relationship, we have $m = EI\frac{d\theta}{ds}$, where $EI$ is the flexural rigidity of the sensing assembly, and the angle $\theta$ is shown in Fig. \ref{fig:modeling_schematic}(c). Substituting the moment-curvature relationship into our moment equilibrium equation, we have:
\begin{align}\label{eq:moment_equilibrium_final}
    \frac{d^2\theta}{ds^2} + (\frac{P}{EI}) \sin{\theta} + (\frac{R}{EI}) \cos{\theta} = 0
\end{align}
the combined flexural rigidity of the sensing assembly is obtained using the mechanical properties and dimensions of the NiTi substrate and the FBG fiber: $EI = E_{fbg} I_{fbg} + E_{wire} I_{wire}$. Here $I_{fbg} = \pi r^2_{fbg}(r^2_{fbg}/4 + \Delta y^2_{na})$ and $I_{wire} = \pi r^2_{wire}(r^2_{wire} + 4 \Delta y^2_{wire})$ are the second moments of area for the FBG fiber and the NiTi wires substrate respectively. Furthermore, the elastic modulus of the FBG fiber, $E_{fbg}$ = 67~GPa, and the Nickel-titanium wires, $E_{wire}$ = 55~GPa, are provided by the manufacturers.
The governing equation (Eq. (\ref{eq:moment_equilibrium_final})) has the following boundary conditions:
\begin{align}\label{eq:boundary_conditions}
    \theta(0) = 0, \quad \frac{d\theta}{ds}(L) &= 0 
\end{align}
The Cartesian coordinates $(x(s), y(s))$ along the arc-length $s$ are defined such that:
\begin{align} \label{eq:cartesian_diff_eq}
    \frac{dx}{ds} = \cos{\theta}, \frac{dy}{ds} = \sin{\theta}
\end{align}
Based on the general geometry of the buckling beam, $x(0)=0$, $y(0)=0$, and $x(L)=0$ hold as trivial conditions in our geometry equations.

To solve for the complete shape of the sensor assembly post-buckling, a numerical solution is approached via a shooting method. The reaction force depicted by $R$ in Eqs. (\ref{eq:moment_balance}) and (\ref{eq:moment_equilibrium_final}) is unknown (similar to P and $\theta$), but in the course of buckling, it varies from zero (when the beam has not buckled at all) to higher values. Since reaction force, $R$, is unknown, dividing Eq. (\ref{eq:moment_equilibrium_final}) by $R$ and substituting $t = s\sqrt{R/EI}$, $\eta = y\sqrt{R/EI}$, and $\kappa = P/R$ into Eq. (\ref{eq:moment_equilibrium_final}), we get the following equation (similar to the approach in \cite{wang1997post}):
\begin{align}\label{eq:variable_change_dq_dt}
    \frac{d^2\theta}{dt^2} + \kappa \sin{\theta} + \cos{\theta} = 0 
\end{align} 
\vspace{-0.2cm}
\begin{align}
    \frac{d\eta}{dt} = \sin\theta \nonumber
\end{align}
Note, that this reformatting allows us to incorporate the unknown $R$ into a variable $\kappa$ that can be assumed to be known and varied to get numerical solutions for Eq. (\ref{eq:moment_equilibrium_final}) for each $\kappa$. This can then be used to compute $R$ for each solution.
Furthermore, the initial conditions are given as follows:

\begin{align}\label{eq:shooting_checks}
    \theta(0) = 0, \quad \frac{d\theta}{dt}(t_{end}) &= 0   
\end{align}
and the final point boundary condition is:
\begin{align}\label{eq:y_boundary_condition}
    \eta(t_{end}) = 0
\end{align}
Note that here, $t_{end}$ is unknown, but for a given (and assumed known) value of $\kappa$ and an initial guess for $\frac{d\theta}{dt}(0)$, we solve Eq. (\ref{eq:moment_equilibrium_final}) until Eq. (\ref{eq:y_boundary_condition}) is satisfied, along with the conditions in Eq. (\ref{eq:shooting_checks}). 
Once all conditions are met, we assume that Eq. (\ref{eq:variable_change_dq_dt}) is solved numerically. This allows us to compute the unknown reaction forces and axial forces, since by our definition, $R = (t^*_{end} / L)^{2}$ and $P = R\kappa$. Here, $t^*_{end}$ is the solution of the shooting function for the dummy variable $t$. Substituting these values into our original (normalized) model Eqs. (\ref{eq:moment_equilibrium_final}) and (\ref{eq:cartesian_diff_eq}), we have a solution for the evolution of $\theta(s)$ and the Cartesian coordinates $(x(s), y(s))$ with respect to arc-length $s$.

\subsection{Finite Element Simulation}
\begin{figure}
\vspace{0.1cm}
\centering\includegraphics[width=\linewidth,keepaspectratio]{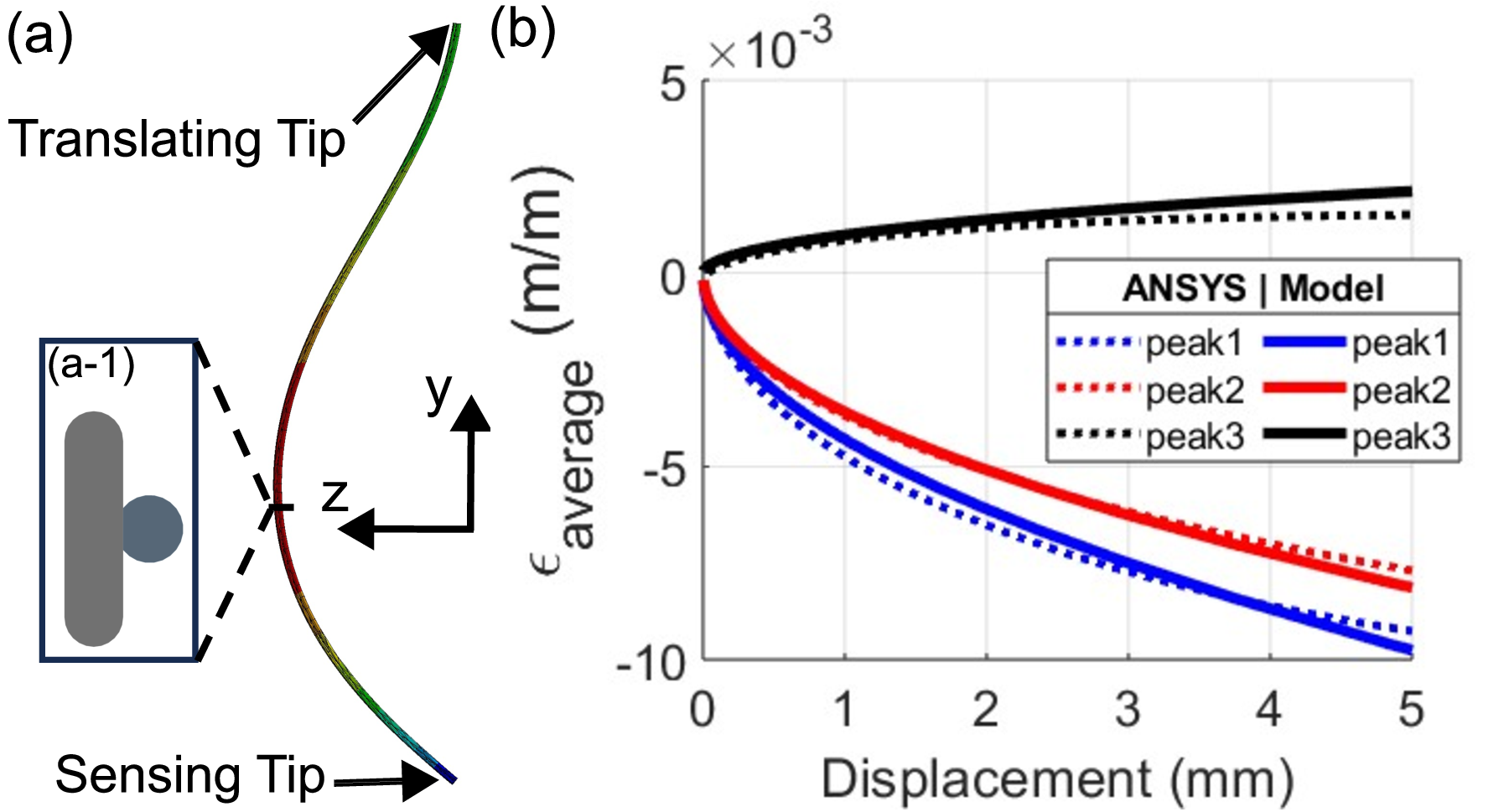}
\caption{(a) Ansys model of stiffness sensor assembly buckling at 5 mm displacement, (a-1) cross-section of the model used to conduct FEA, (b) Comparison between Ansys simulation (dotted) and theoretical model (solid lines) of the axial strain along the gratings/peaks of the FBG fiber.}
\label{fig:ansys}
\vspace{-0.3cm}
\end{figure}
Finite element analysis of the buckling beam is conducted in Ansys Mechanical (Ansys Inc., Pennsylvania, USA) to validate the theoretical model. The sensor is modeled as a 42 mm long beam with a cross-section represented in Fig. \ref{fig:ansys}(a-1) and a Young's Modulus of 66 GPa and a Poisson's Ratio of 0.33. A static structural analysis is used with ``large deflection`` incorporated to allow buckling. The sensing tip displacement is constrained in all three directions, while the rotation is constrained in only the \textit{y}- and \textit{z}-axes (See Fig. \ref{fig:ansys}(a)), thus allowing bending about the $x$-axis. The end of the sensor attached to the linear stage is displaced 5 mm in the axial direction of the sensor and 0.001 mm as an introduced perturbation to allow the simulated model to buckle. The strain values are recorded along the three grating lengths. The described simulation parameters result in the sensor model buckling, as illustrated in Fig. \ref{fig:ansys}(a). The axial strain along each grating length is recorded and compared to the model's, shown in Fig. \ref{fig:ansys}(b). The results show that the theoretical model aligns well with the Ansys simulation.

\subsection{Experimental Validation}

\begin{figure}
\vspace{0.1cm}
\centering\includegraphics[width=\linewidth,keepaspectratio]{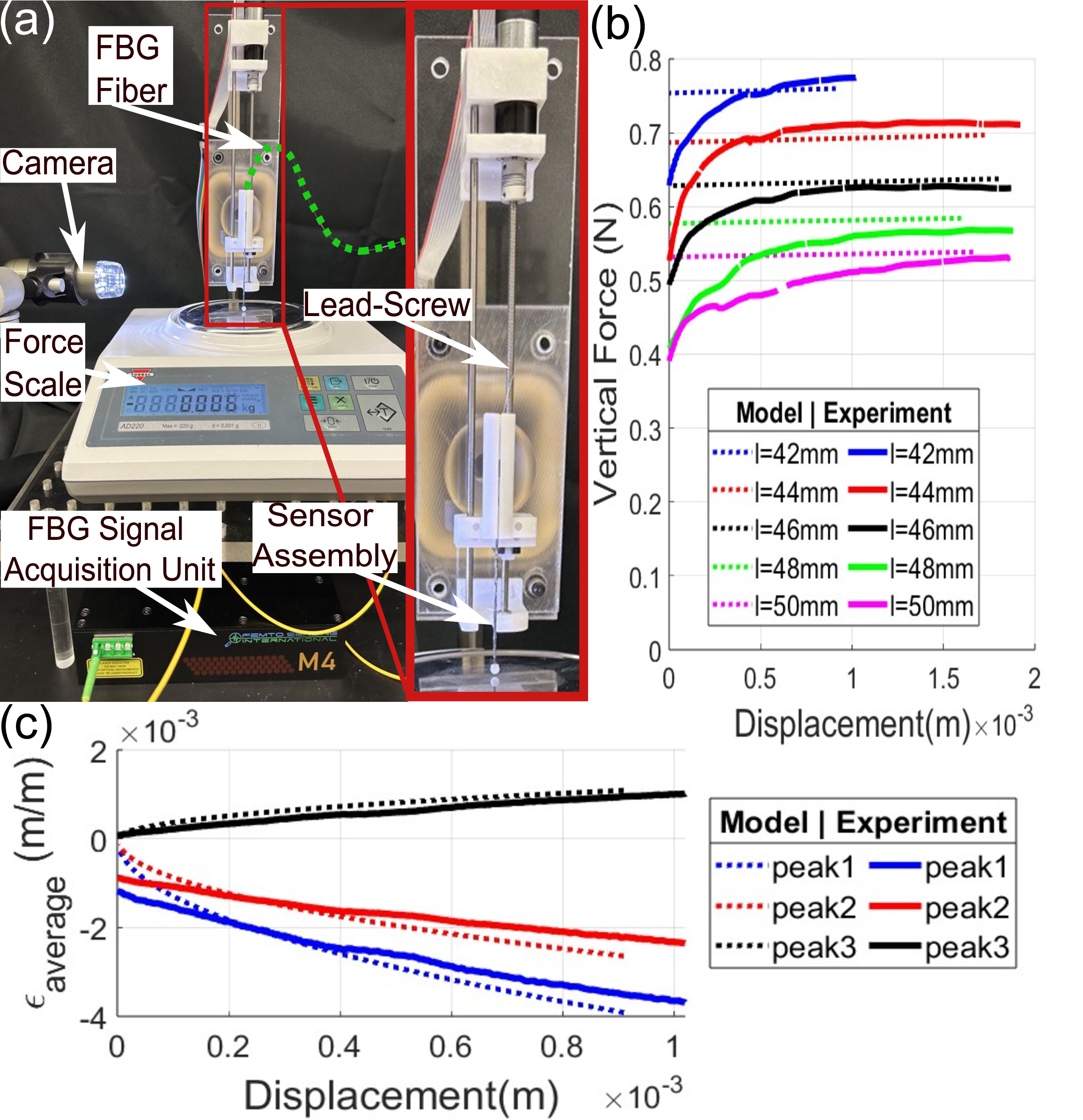}
\caption{(a) Vertical indentation experimental setup used for validating the proposed mechanical model and conducting bench-top tissue sample stiffness estimations, (b) Comparison between post-buckling experimental vertical forces and the corresponding predicted values by the theoretical model for multiple length values, (c) Comparison between the average axial strain of the FBG fiber over the gratings/peaks and the predicted corresponding values by the theoretical model with the length of the sensor being 42mm.}
\label{fig:experimental_validation}
\vspace{-0.5cm}
\end{figure}

To validate the theoretical model developed in Section \ref{subsec:buckling_model} an experimental setup is designed to execute vertical indentation trials on an arbitrary surface. This experimental setup (see Fig. \ref{fig:experimental_validation}(a)) consists of a motor and a lead screw connected to its shaft via a coupling. The sensor assembly is clamped in a 3D-printed nut that can translate vertically on the lead-screw when the motor is actuated, displacing the sensor. The sensor is oriented vertically with its tip pointing downwards. A force scale (Torbal AD220 precision scale, Scientific Industries Inc.) is placed at a suitable height for the sensor to come in contact with its measuring surface when lowered by the motor and lead-screw system. A camera (Dino-Lite digital microscope, Dunwell Tech, Inc.) is pointed at the sensor to capture visual data from the sensor assembly throughout the experimental trials. The FBG optical fiber is routed out from the sensor to its signal acquisition unit (FSI M4 Interrogator, Femto Sensing International, Technica Optical Components, LLC). The Texas Instruments LAUNCHXL-F28379D board is used to collect real-time data from the experimental setup. 

\textcolor{black}{This vertical indentation experimental setup is used to collect vertical force data of the sensor beam as it buckles, FBG fiber strain data (to be translated into the shape of the buckling sensor)}. Upon comparing the collected strain and comparing the same with the values estimated by the model (see Fig. \ref{fig:experimental_validation}(b,c)), the model differs significantly from the data for small post-buckling displacements (i.e., initial post-buckling behavior deviates from the model). However, as the displacement gets higher and the sensor attains higher curvatures, both vertical force and axial strain of the fiber converge more closely to their predicted values. The strain values shown in Fig. \ref{fig:experimental_validation}(c) are average strain over the specific known area where the gratings are located on the fiber. The way this average is replicated in the mathematical model is as follows:
\begin{equation} \label{eq:average_strain}
\epsilon_{average}(s_{1},s_{2}) = \frac{\int_{s_{1}}^{s_{2}} \epsilon(s) \,ds}{|s_{2}-s_{1}|}
\end{equation}
Where $\epsilon_{average}(s_{1},s_{2})$ is the mean value of strain in the length  between $s=s_{1}$ and $s=s_{2}$ on the buckling beam. Note that $\epsilon(s)$ is a continuous function for strain at the arc length $s$, which is obtained using Eq. (\ref{eq:Strain_Curvature_Relation}).

\section{Experiments and Results} \label{sec:experiments_results}
The proposed sensor is used to estimate the stiffness of artificial tissue samples with two types of experiments. First, multiple bench-top trials are conducted to evaluate the sensor's accuracy. After primary assessment, the sensor is installed on a 3D-printed mock concentric tube robot, and the stiffness estimation procedure is conducted on two selected samples.
\subsection{Bench-top Stiffness Estimation} \label{subsec:benchtop_estimations}
\begin{figure} \label{fig:Vertical_estimation}
\vspace{0.1cm}
\centering\includegraphics[width=\linewidth,keepaspectratio]{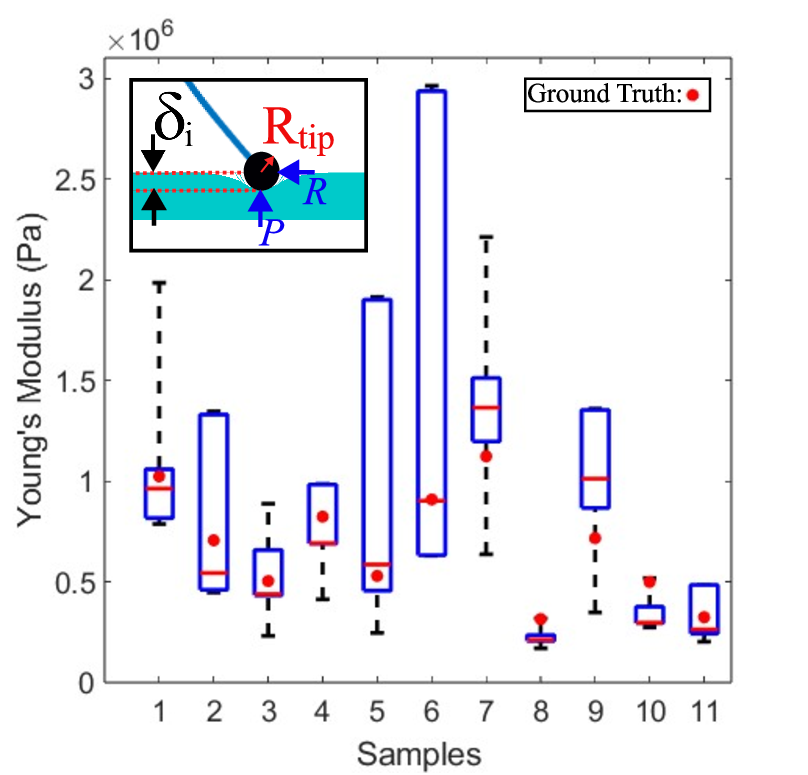}
\caption{Bench-top experimental estimation results and ground truth values. (inset) Close-up schematics of the sensor spherical tip indented into the tissue surface.}
\label{fig:benchtop_estimations}
\vspace{-0.5cm}
\end{figure}
The vertical indentation experimental setup, used in Section \ref{sec:Model} to validate the mechanical model, is utilized to estimate the stiffness of 11 different tissue phantoms made from PDMS and PA-Gel. For each tissue sample, five trials with different sensor lengths of $l_{sensor}=$\{42~mm, 44~mm, 46~mm, 48~mm, 50~mm\} are conducted. \textcolor{black}{The vertical indentation force data, image data, FBG data, and actuator encoder data are gathered and processed.} The results of our stiffness estimation, after removing outliers (above 3~MPa), are shown and compared with ground truth values in Fig. \ref{fig:benchtop_estimations}. To obtain the ground truth values, a Mark-10 ESM 303 (Mark-10 Corporation, NY, USA) equipped with a load cell and a cylindrical 3D-printed indenter tip is used. The ground truth values are calculated using the cylindrical indenter equation in \cite{McKee2011Tissue}.

The elastic modulus is estimated after indentation trials using the following specific equation for spherical indenter \cite{McKee2011Tissue}:
\begin{equation} \label{eq:sphere_indenter}
    E_{t}=\frac{3}{4}\frac{(P(1-\upsilon^2))}{\sqrt{\delta_{i}^3R_{tip}}}
\end{equation}
where $E_{t}$ is the Young's modulus of the surface of the tissue sample \textcolor{black}{that the sensor makes contact with.} Furthermore, $P$ denotes the vertical force \textcolor{black}{experienced by the sensor assembly}, $\upsilon$ is the Poisson's ratio of the tissue sample \textcolor{black}{(which is a known value)} , $\delta_{i}$ shows the amount of indentation \textcolor{black}{made by the spherical tip of the sensor assembly into the surface of the tissue sample}, and $R_{tip}$ is the radius of the spherical part at the tip of the sensor assembly \textcolor{black}{that makes contact with the tissue surface and is} equal to $R_{tip}=3.5~mm$ (See Fig. \ref{fig:benchtop_estimations}(inset)).
The vertical force and image data is used to capture the encoder value corresponding to the moment of contact ($X_{encoder}^{contact}$). Beyond this point, the sphere at the tip of the sensor assembly starts to sink into the surface of the tissue until the sensor assembly begins to buckle. The real-time average strain values\textcolor{black}{, which are computed from the FBG signals, }are observed in this step. As shown in Fig. \ref{fig:experimental_validation}(c), average strain values at different locations on the sensor are expected to diverge when the buckling starts, so a threshold value for the Euclidean norm of the three strain values (at peaks 1 to 3) is considered. Once the strain values pass this threshold, the buckling has commenced and the encoder data is captured only at the buckling point (and indicated by $X_{encoder}^{buckling}$). The parameter $\delta_{i}$, corresponding to the extent of indentation pre-buckling, used in Eq. (\ref{eq:sphere_indenter}), will then be obtained as follows:
\begin{equation} \label{eq:indentation_value}
    \delta_{i} = X_{encoder}^{buckling} - X_{encoder}^{contact}
\end{equation}
After it is confirmed by the real-time strain data from the FBG fiber that the beam is buckled, the exerted force (denoted by $P$) is obtained using Eq. (\ref{eq:moment_equilibrium_final}) and the fixed known length of the sensor. Plugging this vertical force value in Eq. (\ref{eq:sphere_indenter}) along with the obtained $\delta_{i}$, the stiffness of the tissue sample is estimated. 
\begin{table}[]
    \begin{center}
    \caption{Quantitative comparison between the estimated values derived from bench-top experimental trials and the authentic stiffness values of the samples.}
        \begin{NiceTabular}{m{1cm} m{1.5cm} m{1.5cm} m{1.2cm} m{1.2cm}}[hvlines,corners=NW] 
             Sample &  $E_{t}^{estimated}$ (KPa) & $E_{t}^{actual}$ (KPa) & RMSE (KPa) &  IQR (KPa)\\
             1 & 963.23 & 1023.67 & 465.46 & 239.62 \\
             2 & 543.36 & 706.30 & 393.00 & 870.74 \\
             3 & 438.82 & 504.03 & 223.95 & 225.30 \\
             4 & 692.04 & 823.77 & 236.16 & 293.40 \\
             5 & 585.91 & 529.33 & 714.43 & 1445.29 \\
             6 & 902.28 & 908.88 & 1150.19 & 2306.26 \\
             7 & 1365.47 & 1122.86 & 582.93 & 178.92 \\
             8 & 211.94 & 314.53 & 93.87 & 29.12 \\
             9 & 1012.38 & 716.74 & 411.31 & 486.53 \\
             10 & 296.86 & 499.56 & 165.90 & 82.00 \\
             11 & 263.03 & 323.78 & 115.30 & 241.83 \\
        \end{NiceTabular}
        \label{tab:E_comparison_bench_top}
    \end{center}
    \vspace{-0.7cm}
\end{table}
As it is shown in Fig. \ref{fig:benchtop_estimations} and Table \ref{tab:E_comparison_bench_top}, the utilized method predicts the actual stiffness in most cases (the average reported RMSE value across all the tissue samples is 413.86 KPa).
\subsection{Stiffness Estimation with a Robot}
\begin{figure}
\vspace{0.1cm}
\centering\includegraphics[width=\linewidth,keepaspectratio]{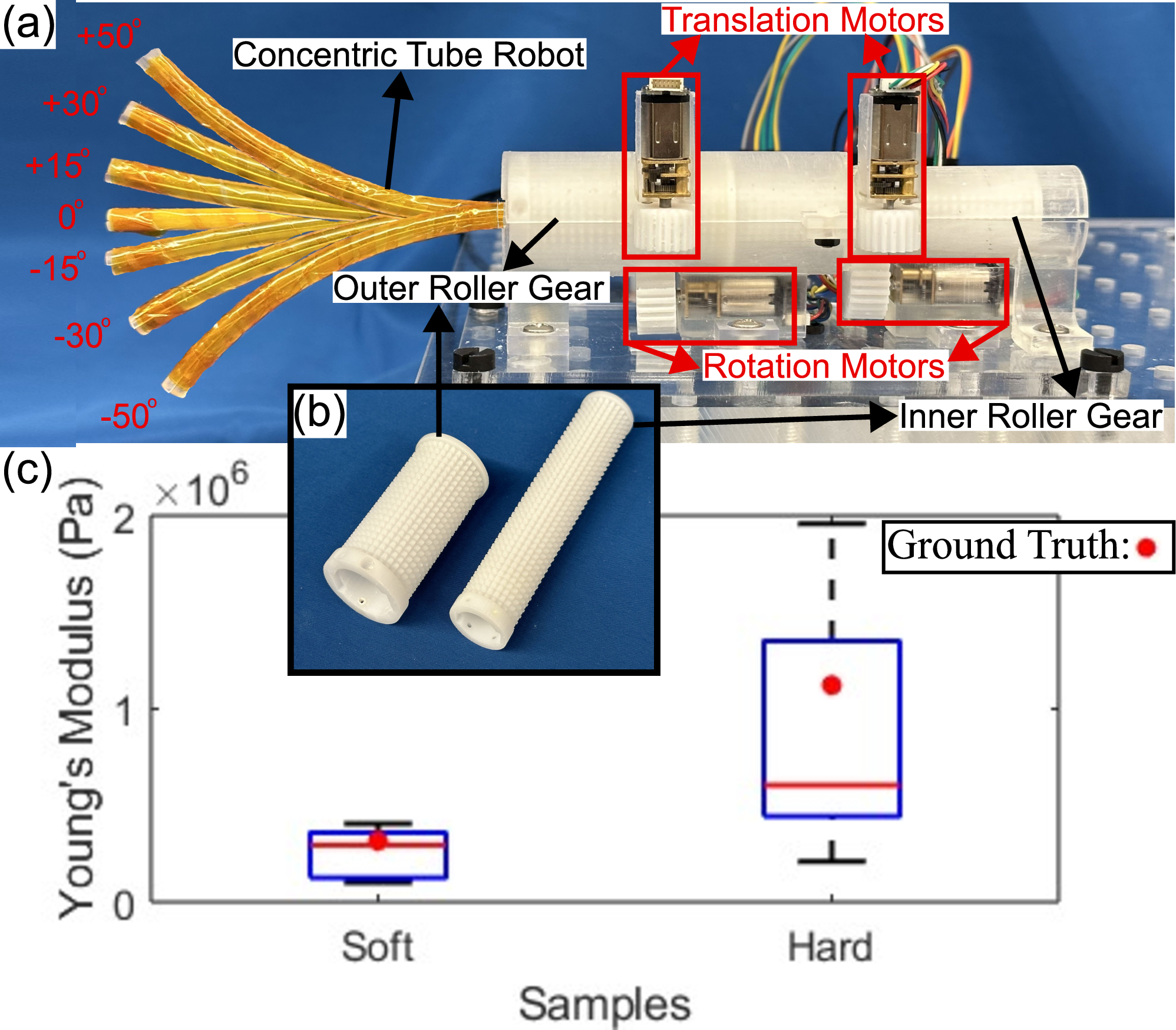}
\caption{(a) Robotic setup consisting of the dual roller gear mechanism and multiple 3D-printed mock concentric tube robots with tip angles of 0, 15, 30, and 50 degrees pointing in two (downwards and upwards) directions, (b) Outer and inner roller gears situated external to the actuation mechanism assembly, (c) Stiffness estimation results for soft and hard tissue compared with their corresponding actual values.}
\label{fig:robot}
\vspace{-0.3cm}
\end{figure}
After conducting the bench-top experiments, the proposed stiffness sensor is installed on a 3D-printed mock concentric tube robot for non-vertical indentation trials. Therefore, an experimental setup is developed and assembled to navigate the sensor suitably to conduct perpendicular indentations over the targeted tissue surface. In this experimental setup, a pair of roller gears (namely outer and inner roller gears, see Fig. \ref{fig:robot}(b)) are used. These roller gears are designed to obtain simultaneous translational and rotational actuation (inspired by \cite{morimoto2017design}). 

Multiple tubes with different tip angles (including 0, 15, 30, and 50 degrees) and constant curvature center-lines are also 3D-printed and attached to the outer roller gear which can only be rotated (See Fig. \ref{fig:robot}(a)). The sensor assembly is fixed at the distal part of a push-rod attached to the inner roller gear. The push rod, holding the sensor at the tip, is placed inside the 3D-printed tube. The inner roller gear is not only programmed to rotate with the outer roller gear (so that twisting between the robot and the sensor will not occur), but it can also be translated to advance/retract the sensor for indentation tasks once the tip is at a suitable distance from the tissue surface and the advancement direction of the sensor axis is orthogonal to the surface of the tissue. The optical fiber is also routed through the tube and outer roller gear as well as at the back of the inner roller gear, going to the FBG signal acquisition unit shown in Fig. \ref{fig:experimental_validation}(a).
Two tissue samples are selected from the batch used in Section \ref{subsec:benchtop_estimations} that will be referred to as soft and hard samples in this section. These tissue samples are fixed in multiple places with suitable heights and orientations for the robot to conduct indentation trials. Only one indentation trial is conducted on each sample using the straight tube. For all the other tubes, three different trials (in case of the position of the tissue sample placement) are conducted on each tissue sample: 1. downward, 2. sideways, and 3. upward.

\textcolor{black}{In each trial, encoder data from the motor for advancing the sensor out of the tube, the average strain data from the FBG signals, and the image data are gathered.} Using the encoder data and the initial exposed length of the sensor, the length of the sensor assembly is recorded in the course of each experiment. The image data captures the encoder value corresponding to the moment of contact. The strain data are also observed to find the encoder value corresponding to the moment that buckling starts in the same manner as explained in Section \ref{subsec:benchtop_estimations}. Subsequently, Eq. (\ref{eq:indentation_value}) is used to obtain the amount of indentation. Based on the recorded encoder value at the moment of buckling, the exposed length of the sensor is derived and the vertical exerted force can be calculated using the mechanical model. This force value, along with the amount of indentation, is then substituted into Eq. (\ref{eq:sphere_indenter}) to estimate the stiffness of the tissue sample.
\begin{table}[]
    \begin{center}
    \caption{Quantitative analysis of estimated values acquired through indentation trials employing the mock concentric tube robot, compared with the actual stiffness values of the samples.}
        \begin{NiceTabular}{m{1cm} m{1.5cm} m{1.5cm} m{1.2cm} m{1.2cm}}[hvlines,corners=NW] 
             Sample &  $E_{t}^{estimated}$ (KPa) & $E_{t}^{actual}$ (KPa) & RMSE (KPa) &  IQR (KPa)\\
             Soft & 602.95 & 1122.86 & 649.27 & 909.27 \\
             Hard & 293.09 & 314.53 & 119.95 & 233.23 \\
        \end{NiceTabular}
        \label{tab:E_comparison_robot}
    \end{center}
    \vspace{-0.7cm}
\end{table}
The estimated values are shown in Fig. \ref{fig:robot}(c) and compared with the corresponding ground truth values. The estimated values are quantitatively compared with the actual value of the samples' Young's moduli, and their corresponding RMSE values are reported in Table \ref{tab:E_comparison_robot}.

\subsection{Failure Cases}
As shown in Fig. \ref{fig:benchtop_estimations}, the stiffness ranges measured by our sensor are relatively high for some tissue samples (particularly for samples 5 and 6). The notable variability in IQR values is likely attributed to inherent differences in sample physical properties. Factors such as surface friction and sample thickness may be key contributors. Further exploration is needed to clarify their specific impact on the observed outcomes. However, the samples used in this study were in a higher stiffness range compared to the real tissue samples (e.g. skin tissue is 6~KPa, and brain is 1.8~KPa). Therefore, accounting for this and adjusting our sensor assembly to lower buckling forces by reducing its rigidity is crucial for future studies. Similarly, there's a comparable issue in estimating the stiffness of the hard sample shown in Fig. \ref{fig:robot}(c), likely for the same reasons. Moreover, the calculated Young's modulus for the hard sample is significantly lower than its actual stiffness. This discrepancy is likely due to small movements in the tip of the mock concentric tube robot during indentation, caused by the low bending rigidity of the 3D-printed tube compared to the exerted force by sensor assembly. Another major limitation of this work is the uncertainty in predicting point-of-contact with tissue, which affects the accuracy of the method, making it more challenging when the targeted tissue is relatively stiff. 
\section{Conclusions} \label{sec:conclusions}
In conclusion, our study indicates that the shape of a beam buckling upon indentation on the surface of a tissue may be used to derive material properties of the surface being indented. A beam specifically manufactured using an FBG fiber and four nitinol rods was designed capable of measuring its own shape when it undergoes buckling. A mathematical model was developed for the fixed-pinned beam's buckling behavior to estimate the strains across grating elements in the beam. The model shows a promising agreement with both finite element simulation and experimental data, and bench-top estimation results align well with the actual stiffness of the samples tested. The proposed sensor was then installed on a 3D-printed concentric tube robot and showed successful differentiating between soft and hard tissue samples. In our future work, we will address one major limitation of this work, i.e. estimating the point of contact accurately, using another highly sensitive grating at the very tip of the FBG fiber.
\bibliographystyle{IEEEtran}
\bibliography{references}
\end{document}